\pdfoutput=1

\documentclass[switch]{article}

    \usepackage[preprint]{neurips_2024}

\usepackage[utf8]{inputenc} 
\usepackage[T1]{fontenc}    
\usepackage{hyperref}       
\usepackage{url}            
\usepackage{booktabs}       
\usepackage{amsfonts}       
\usepackage{nicefrac}       
\usepackage{microtype}      
\usepackage{xcolor}         

\usepackage{amsmath}
\usepackage{bm}

\usepackage{graphicx}       

\usepackage{multirow}
\usepackage{multicol}
\usepackage{array}
\usepackage{rotating} 
\usepackage{tabularx} 
\usepackage{makecell} 

\usepackage{setspace} 

\usepackage{tcolorbox} 

\usepackage{utfsym} 

\usepackage{ulem} 

\usepackage{pgfplots} 

\usepackage{enumitem} 

\usepackage{caption} 

\usepackage{algorithm}      
\usepackage{algorithmic}    


\usepackage{natbib}
\setcitestyle{authoryear,round,citesep={;},aysep={,},yysep={;}}

\definecolor{lzqPurple}{RGB}{148, 39, 212}
\definecolor{lightRed}{RGB}{253, 190, 172}
\definecolor{lightGreen}{RGB}{172, 253, 187}
\definecolor{lightYellow}{RGB}{253, 248, 172}

\title{SED: Self-Evaluation Decoding Enhances \\
Large Language Models for Better Generation}

\author {
    Ziqin Luo\textsuperscript{\rm 1},
    Haixia Han\textsuperscript{\rm 3},
    Haokun Zhao\textsuperscript{\rm 2},
    Guochao Jiang\textsuperscript{\rm 1},
    Chengyu Du\textsuperscript{\rm 2}, \\
    \textbf{Tingyun Li}\textsuperscript{\rm 2},
    \textbf{Jiaqing Liang}\textsuperscript{\rm 1}\thanks{Corresponding Author},
    \textbf{Deqing Yang}\textsuperscript{\rm 1},
    \textbf{Yanghua Xiao}\textsuperscript{\rm 2,3}
    \\
    Shanghai Key Laboratory of Data Science, School of Data Science, Fudan University \textsuperscript{\rm 1}\\
    Shanghai Key Laboratory of Data Science, School of Computer Science, Fudan University \textsuperscript{\rm 2}\\
    Shanghai Institute of Artificial Intelligence for Education, East China Normal University \textsuperscript{\rm 3}\\
   \texttt{\{zqluo22,gcjiang22\}@m.fudan.edu.cn}, \\
    \texttt{\{haixiahan03,cydu2024,litinyun0715\}@gmail.com}, \\
    \texttt{\{liangjiaqing,yangdeqing,shawyh\}@fudan.edu.cn}
}

\begin{document}

\maketitle


\begin{abstract}
\label{sec:abstract}
Existing Large Language Models~(LLMs) generate text through unidirectional autoregressive decoding methods to respond to various user queries.
These methods tend to consider token selection in a simple sequential manner, making it easy to fall into suboptimal options when encountering uncertain tokens, referred to as chaotic points in our work.
Many chaotic points exist in texts generated by LLMs, and they often significantly affect the quality of subsequently generated tokens, which can interfere with LLMs' generation. 
This paper proposes \textbf{S}elf-\textbf{E}valuation \textbf{D}ecoding, \textbf{SED}, a decoding method for enhancing model generation.
Analogous to the human decision-making process, SED integrates speculation and evaluation steps into the decoding process, allowing LLMs to make more careful decisions and thus optimize token selection at chaotic points.
Experimental results across various tasks using different LLMs demonstrate SED’s effectiveness.

\end{abstract}
\begin{figure*}[h]
\centering 
\captionsetup{aboveskip=7pt, belowskip=0pt} 
\includegraphics[width=1.0\linewidth,page=1]{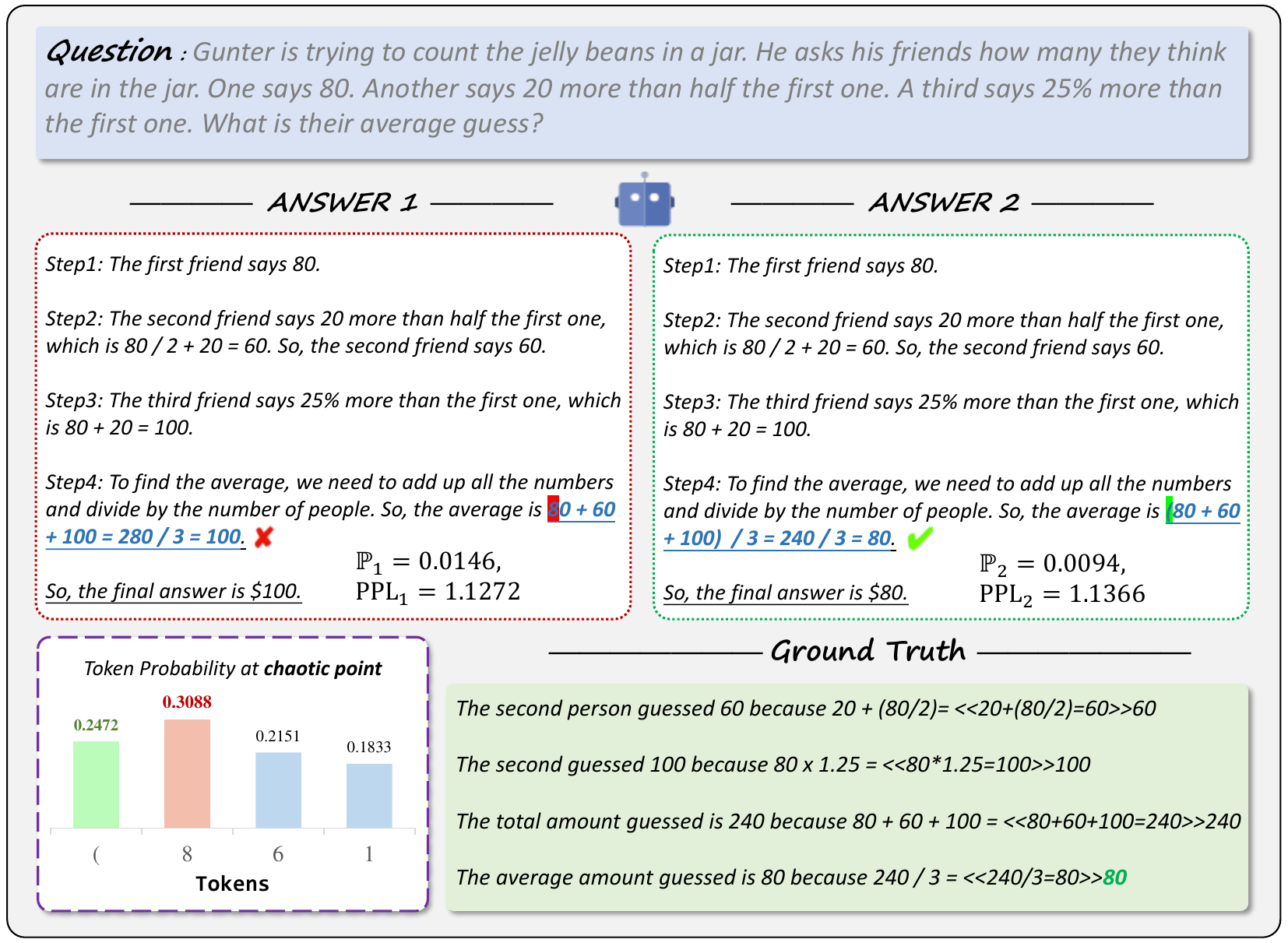}
\vspace{-0.3cm}
\caption{The same model with different decoding methods generates two answers to the GSM8K question. 
The left is generated through greedy search, while the right is produced via SED. 
Tokens with \raisebox{0.5ex}{\colorbox{red}{ }} or \raisebox{0.4ex}{\colorbox{green}{ }} colored are the first chaotic points the model meets.
Underlines represent the final generated text after the chaotic point. 
Corresponding sentence-level probabilities and perplexities are shown next to them.
The token probability mass at the chaotic point is shown in the left-bottom.
}
\label{fig:teaser}
\end{figure*}

\section{Introduction}
\label{sec:introduction}
Large Language Models (LLMs) are equipped with rich word knowledge, relying on decoding methods to express their abilities during the inference phase.
Existing decoding methods are mainly autoregressive and are primarily classified into two categories~\citep{19-decoding-methods-comparison}. Firstly, search-based methods aim to maximize the probability of generated text, such as greedy search and beam search. The other is sampling-based methods, which sample the next generated token based on a probability distribution, such as nuclear sampling.

Despite good results, these autoregressive decoding methods for sequential token generation have limitations.
LLMs often encounter highly uncertain tokens, referred to as chaotic points in this paper, during decoding.
Figure~\ref{fig:teaser} shows the token probability distribution at a chaotic point. 
The probability mass is quite dispersed, which suggests that the model does not know which token to choose.
On the other hand, Answer 1 and Answer 2 in Figure~\ref{fig:teaser} have the same content before the chaotic point but with different conclusions.
At the chaotic point, Answer 1 selects token `8', resulting in an incorrect response, whereas Answer 2 chooses the token `(', leading to a correct one.
Different token choices lead to vastly different results, indicating that tokens at chaotic positions should be carefully selected.
However, greedy search only considers the token with the highest probability, ignoring other tokens with considerable probability mass. 
Sampling is easily influenced by randomness and falling into irrelevant tokens.
Though beam search can try multiple tokens, it only considers sentence-level probability maxima. 
In Figure~\ref{fig:teaser}, Answer 1 has a larger sentence-level probability and a lower perplexity than Answer 2, so beam search still fails to respond correctly.

The case in Figure~\ref{fig:teaser} implies that the model's incorrect answer may not be due to its lack of relevant knowledge but because it does not know how to choose an optimal token at the chaotic point. 
LLMs should not only take the probabilities into account when selecting the token at a chaotic point.
However, existing autoregressive decoding methods are limited to probability, not considering selecting the token from a holistic perspective. 
This could easily lead to inappropriate tokens and thus result in error propagation.
Therefore, it is necessary to propose a more advanced decoding method to improve LLMs' token selection at chaotic points.

When making uncertain decisions, humans often speculate first and then evaluate the speculations, compare them, and make the final choice~\citep{79:prospect-theory}, a common strategy called slow thinking~\citep{17-thinking-fast-and-slow} people use when dealing with complex decisions.
This approach can assist LLMs in selecting appropriate tokens at chaotic points, but it requires LLMs to be equipped with evaluation ability.
However, \cite{23-critique-ability-llm} suggest the inherent evaluation ability of existing LLMs is weaker than their generation ability.
Recently, many studies~\citep{23-autoj,23-judgelm,23-critiquellm} have shown that LLMs can significantly enhance their evaluation ability and thus achieve self-correction~\citep{24-self-correct} through training.
In this paper, we propose \textbf{S}elf-\textbf{E}valuation \textbf{D}ecoding, \textbf{SED}, a decoding method for improving uncertain tokens selection at chaotic points for LLMs. 
Mirroring human decision-making, when encountering a chaotic point during decoding, LLMs first speculate the results of choosing each token, then evaluate them, and finally select the token with the highest propensity score based on evaluations.

Our main contributions are summarized as follows:
\begin{itemize}[leftmargin=0.5cm, topsep=-2pt] 
\item We propose straightforward criteria that help LLMs efficiently identify chaotic points and thus carefully select key tokens during decoding.
\item We propose Self-Evaluation Decoding (SED). Analogous to human decision-making, SED enables LLMs to speculate and evaluate uncertain tokens at chaotic points to select the optimal token and thus improve the generation's quality compared to other standard decoding methods.
\item We design a simple but effective data synthesis strategy for enhancing LLMs' evaluation ability based on multi-model ensembling, achieving self-evaluation in SED. 
\end{itemize}

\section{Related work}

\textbf{Enhancing evaluation ability for LLMs.}
Recent studies~\citep{23-gptscore,23-critique-ability-llm,23-llm-as-a-judge} reveal that LLMs' evaluation ability can be seen as the ability to assess the quality of the object being evaluated (usually a question-answer pair).
This ability is determined by the accuracy of the feedback generated by the evaluation model denoted as $\bm{M}_\text{eval}$. 
The feedback is usually one rating (scalar), analysis (natural language text), or a combination of both. 
The more accurate the feedback, the better the evaluation. 
The approaches to enhancing LLM's evaluation ability can be mainly categorized into \textit{prompt-based} and \textit{tuning-based}. 
The former designs well-structured prompts with detailed instructions to guide the model to evaluate. 
Since this approach relies entirely on the model's inherent capability, closed-source models such as GPT-4~\citep{23-gpt4} and PaLM 2~\citep{23-palm2} are usually adopted as evaluation models, making the evaluation more expensive. 
However, being untrained, this method is difficult to adapt to specific downstream tasks and is greatly limited by the model's inherent flaws~\citep{23-not-fair-evaluators,23-critique-ability-llm}. 
The latter builds training datasets consisting of evaluation feedback to inject evaluation abilities into open-source LLMs through finetuning explicitly~\citep{23-pandalm,23-shepherd}. 
Many studies have demonstrated that the evaluation models built on direct training can achieve near-human evaluation quality in dialog, open-scene QA, fact verification, and other scenarios. 
These are effective and low-cost solutions for building evaluation models.
In this work, we organically combine the LLMs' evaluation ability with the other inherent ability, so that LLMs can not only generate but also evaluate the text it generates.
Specifically, we enhance the model the evaluation ability through training over the augmented dataset with evaluation feedback included. 
We will talk about the details in Section~\ref{sec:eval-abi-cons}.


\textbf{Decoding methods for LLMs.}
Recent studies~\citep{23-llm-survey,24-unlocking} have focused on how to improve decoding methods for LLMs further. 
They can be partitioned into two types: \textit{acceleration-oriented} and \textit{quality-oriented}. 
The former focuses on accelerating the generation of LLMs from an algorithmic perspective. 
One representative work is speculative decoding~\citep{24-unlocking}, whose core idea is to use lightweight models with much lower decoding overhead for generation, and heavy-weight models to verify the generated tokens. 
In this way, the parallelism of decoding substantially increases thus realizing acceleration. 
The latter focuses on enhancing the quality of LLMs' generation~\citep{23-dola,24-in-context,24-flap}. 
The main operation is to adjust the unnormalized logits during decoding in light of some signals or criteria. 
For instance, \cite{22-contrastive-decoding} enhances generation diversity by contrasting the logits of the stronger model over the weaker one. 
\cite{24-computer-parsable} guides the model to sample correct tokens in the decoding phase by introducing constraints to make the model output more parsable to the computer.
\cite{24-deal} acquires reward signals based on formatted programs or parametric models and then makes the decoding more controllable. 
In contrast, we utilize signals from the model itself to guide decoding, thereby leveraging LLMs' evaluation ability to improve the generation.

\begin{figure*}[h]
\centering 
\captionsetup{aboveskip=7pt, belowskip=-10pt} 
\includegraphics[width=1.0\linewidth,page=1]{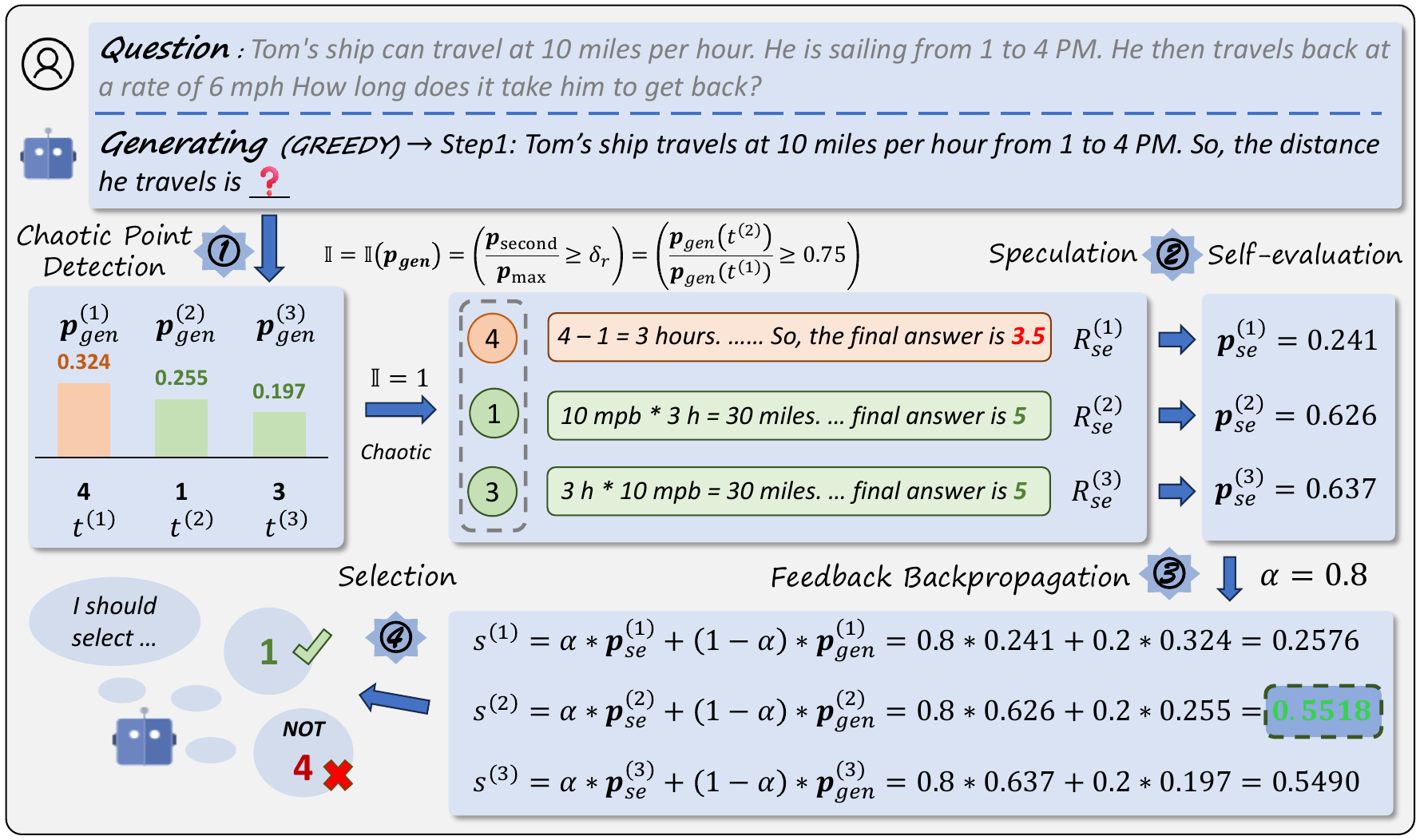}
\vspace{-0.3cm}
\caption{The workflow  of token selection at \textbf{ONE} chaotic point during decoding. 
Other chaotic points also follow a similar process.
In this example, we adopt the ratio-based detection criterion and consider top $3$ tokens for speculation and evaluation. 
We set $\alpha=0.8$ and $\delta_r=0.75$.
}
\label{fig:framework}
\end{figure*}

\section{Self-evaluation decoding}
\label{sec:method}



While previous standard left-to-right decoding methods treat all tokens equally and employ a uniform policy, our proposed self-evaluation decoding (SED) emphasizes the tokens generated at chaotic points, yielding better responses. 
This paper primarily addresses three critical questions: 
(1) How can a chaotic point be identified? 
(2) Which tokens could be considered at the chaotic point? 
(3) Which token is the most suitable choice at the chaotic point? 
To this end, our SED method encompasses three modules: $\textit{chaotic point detection}$, $\textit{speculation and evaluation}$, and $\textit{feedback backpropagation}$. 

Autoregressive decoding generates tokens sequentially.
Specifically, consider a sequence of generated tokens $\{t_1,t_2,\dots,t_{n}\}$, where $n$ represents the total number of tokens.
Each token $t_i$(with $i\in[1,n]$) is generated from the distribution $\mathbb{P}_i=P(\cdot|x,t_{<i})$, which is conditioned on the input $x$ and the preceding tokens $t_{<i}=\{t_1,t_2,\dots,t_{i-1}\}$, over the whole vocabulary $\mathcal{V}$.
During decoding, SED examines the conditional distribution $\mathbb{P}_i$ and calculates the indicator $\mathbb{I}_i:=\mathbb{I}(\mathbb{P}_i)$ to determine whether the token $t_i$ is hard for the model to predict. 
If $\mathbb{I}(\mathbb{P}_i)=0$, $t_i$ is selected through default decoding method (e.g. greedy search).
If $\mathbb{I}(\mathbb{P}_i)=1$, the position $\mathfrak{i}$ is identified as a chaotic point. 
SED selects the top $k$ tokens from $\mathbb{P}_i$ for speculation and allows the model to evaluate these speculations. 
The model's self-evaluation is backpropagated and combined with $\mathbb{P}_i$ to guide the selection of the optimal token at position $\mathfrak{i}$.  
By iteratively repeating this process, LLMs generate improved responses.
The full SED pipeline is shown in Algorithm~\ref{alg:self-eval}.
Figure~\ref{fig:framework} shows what happens at a chaotic point in one iteration.
Further details will be elaborated in subsequent sections.

\begin{algorithm}[!h] 
    \caption{ Self-Evaluation Decoding}
    \label{alg:self-eval}
    \renewcommand{\algorithmicrequire}{\textbf{Input:}}
    \renewcommand{\algorithmicensure}{\textbf{Output:}}
    \begin{algorithmic}[1]
        \REQUIRE user query $\mathcal{Q}$, 
        model $\mathcal{M}$, 
        tokenizer $\mathcal{T}$, 
        max new tokens $m$, 
        threshold $\delta$, 
        number of token candidates $k$,
        vocabulary $\mathcal{V}$
        \ENSURE response $\mathcal{R}$ to the user query   
        
        \STATE  \texttt{generated\_tokens} = [] 
        
        \FOR{each $i \in [1,m]$}
            \STATE $\mathbb{P}_i \leftarrow$  \textsc{cal\_probs}$(\mathcal{M},\mathcal{Q},\texttt{generated\_tokens})$ 
            \quad
            \textcolor{gray}{
                \small\textit{\# probability distribution for predicting $t_i$}
                }
            \STATE $\mathbb{I}_i$ $\leftarrow$ \textsc{is\_chaotic}$(\mathbb{P}_i, \delta)$
            \quad
            \textcolor{gray}{
                \small\textit{\# judge whether position $i$ is a chaotic point}
                }
            \IF { $\mathbb{I}_i == 1$ }
                \STATE \texttt{tk\_cands} $\leftarrow$ \textsc{select\_token\_candidates}$(k, \mathbb{P}_i, \mathcal{V})$
                
                \STATE \texttt{self\_eval\_prob} $\leftarrow$ \textsc{speculation\_evluation}(\texttt{tk\_cands}, $\mathcal{M}$, $\mathcal{V}$)
                
                \STATE \texttt{scores} $\leftarrow$ \textsc{backpropagation}(\texttt{self\_eval\_prob}, $\mathbb{P}$) 
                
                \STATE $t_i \leftarrow$ \textsc{select\_token}(\texttt{scores}, $\mathcal{V}$) 
                \quad\textcolor{gray}{
                \small\textit{\# choose the token with the highest score as $t_i$}
                }
            \ELSE 
                \STATE $t_i \leftarrow$ \textsc{decoding}($\mathcal{M}$, $\mathbb{P}$) \quad\textcolor{gray}{
                \small\textit{\# select $t_i$ by standard decoding method, e.g. greedy, beam, sampling}
                }
            \ENDIF
            \STATE \texttt{generated\_tokens.append}$(t_i)$
            \quad\textcolor{gray}{
                \small\textit{\# record $t_i$ as the generated tokens for subsequent generation}
                }
            \IF{stopping criteria satisfied}
                \STATE \texttt{break}
            \ENDIF
        \ENDFOR
        
        \RETURN \textsc{convert\_tokens\_to\_text}($\mathcal{T}$, \texttt{generated\_tokens}) 
    \end{algorithmic}
\end{algorithm} 

\subsection{Chaotic point detection}
\label{sec:chaotic-point-detection}

\textbf{Entropy-based.} Entropy is commonly used to describe the complexity or uncertainty in information theory. 
During autoregressive decoding, the model generates the probability distribution $\mathbb{P}_i$ over vocabulary $\mathcal{V}$, with the entropy defined as 
\[
    \mathbb{E}_i := \mathbb{E}(\mathbb{P}_i)=-\sum_{j=1}^{\left|\mathcal{V}\right|}\mathbb{P}_i\left(t^{(j)}|x,t_{<i}\right)\log\mathbb{P}_i\left(t^{(j)}|x,t_{<i}\right)
\]
$t^{(j)}$ is the $j$-th token in $\mathcal{V}$. 
We can use $\mathbb{E}_i$ to measure the model's uncertainty at position $\mathfrak{i}$.

\textbf{Ratio-based.} Another more intuitive measurement for uncertainty is the probability ratio:
\[
    \mathbb{R}_i:=\mathbb{R}(\mathbb{P}_i)=\frac{p_\text{sec}}{p_\text{max}}, \quad p_\text{sec}=\mathbb{P}_i(t_\text{sec}|x,t_{<i}), p_\text{max}=\mathbb{P}_i(t_\text{max}|x,t_{<i})
\]
$p_\text{max}$ and $p_\text{sec}$ are the maximum and second-largest probabilities in $\mathbb{P}_i(\dot|x,t_{<i})$, with $t_\text{max}$ and $t_\text{sec}$ being the corresponding tokens.
It follows the clear idea that the model is at least unsure which token between $t_\text{max}$ and $t_\text{sec}$ to take at position $\mathfrak{i}$ when $p_\text{sec}$ is close to $p_\text{max}$.

After obtaining $\mathbb{E}_i$ and $\mathbb{R}_i$, we can determine whether position $\mathfrak{i}$ is a chaotic point.
Define indicator $\mathbb{I}_i:=\mathbb{I}(\mathbb{P}_i)$ as
\[
\mathbb{I}(\mathbb{P}_i) =  \left\{\begin{matrix}
1 & \mathbb{E}_i \ge \delta_e \\
0 & \text{otherwise.} \\
\end{matrix}\right.
\quad\text{ or }\quad
\mathbb{I}(\mathbb{P}_i) =  \left\{\begin{matrix}
1 & \mathbb{R}_i \ge \delta_r \text{ or } p_\text{max}\le(1-\delta_r) \\
0 & \text{otherwise.} \\
\end{matrix}\right.
\]
$\delta_e$ and $\delta_r$ are the uncertainty thresholds for entropy-based and ratio-based detection, respectively.
Position $\mathfrak{i}$ is considered a chaotic point for the model if $\mathbb{I}(\mathbb{P}_i)=1$.
Figure~\ref{fig:framework} intuitively demonstrates how this operates.
Note that, we set a branching factor $b$ as the upper limit for the number of detected points to control the algorithm's complexity.

\subsection{Speculation and evaluation}
\label{sec:speculation-and-evaluation}
If position $\mathfrak{i}$ is chaotic, we need to speculate first.
Specifically, we select top $k$ tokens 
$\{t^{(j)}_i\}_{j=1}^{k}$ 
with the highest probabilities from distribution $\mathbb{P}_i$.
These tokens are appended to the existing context $t_{<i}$ to form $k$ new contexts $\{[t_{<i};t^{(j)}_i]\}_{j=1}^{k}$, differing only in the last token $t^{(j)}_i$. 
The model then completes these $k$ contexts, resulting in $k$ responses $\{R_{se}^{(j)}\}_{j=1}^{k}$.

For evaluation, the model itself evaluates $\{R_{se}^{(j)}\}_{j=1}^{k}$.
As we focus on the QA scenario, the evaluation feedback specifically refers to the likelihood of responses being considered correct. 
Formally, given the primal user question $Q$ and $R_{se}^{(j)}$, the model judge $R_{se}^{(j)}$'s correctness as $J_{se}^{(j)}$. 
we extract the model's output \texttt{logits} and calculate the self-evaluation probability
\[p_{se}^{(j)} := \mathbb{P}\left(J_{se}^{(j)}=1|Q, R_{se}^{(j)}\right)\in[0,1],\quad\text{with }J_{se}^{(j)} = \left\{\begin{matrix}
1 & R_{se}^{(j)}\text{ is correct to }Q \\
0 &  \text{otherwise}\\
\end{matrix}\right.\]
Through $p_{se}^{(j)}$, we can tell the model confidence on the speculation $R_{se}^{(j)}$ and further determine the token selection propensity at position $\mathfrak{i}$.
Considering the inference cost, the model adopts standard decoding methods during speculation. 
Details about $p_{se}^{(j)}$ are presented in Appendix~\ref{sec:appendix-judgement-inference}.
We will discuss the effects of different methods on speculation in Section~\ref{sec:analysis-speculation-alignment}.

\subsection{Feedback backpropagation}
\label{sec:feedback-backpropagation}

After $\textit{Speculation and Evaluation}$, we get $\{p_{se}^{(j)}\}_{j=1}^{k}$ for $k$ potential candidates $\{t^{(j)}_i\}_{j=1}^{k}$. 
Back to position $\mathfrak{i}$, we fuse 
$\{p_{se}^{(j)}\}_{j=1}^{k}$ 
with the token probabilities 
$\{p_{gen}^{(j)}\}_{j=1}^{k}:=\{\mathbb{P}_i(t^{(j)}_i|x,t_{<i})\}_{j=1}^{k}$ 
which are what we already get during decoding. 
$p_{se}^{(j)}$ can be viewed as the model's expected benefit from choosing $t_i^{(j)}$, while $p_{gen}^{(j)}$ represents the model's confidence in choosing $t_i^{(j)}$ under the current conditions. 
$p_{se}^{(j)}$ focuses on the future, whereas $p_{gen}^{(j)}$ focuses on the present. 
We introduce the concept of propensity score by combining the two to assist the model in choosing the optimal token at the chaotic point.
Formally, for $t_i^{(j)}$, we define its propensity score as
\[s_i^{(j)} = \alpha * p_{se}^{(j)} + (1-\alpha) * p_{gen}^{(j)}\]
Here, $\alpha$ is the fusion coefficient controlling the fusion between the evaluation feedback and token probability. The higher $s_i^{(j)}$, the better bidirectional feature token $t^{(j)}_i$ has. Finally, we select the token with the highest propensity score for the subsequent generation.

\subsection{Evaluation ability enhancing}
\label{sec:eval-abi-cons}
\textit{Speculation and Evaluation} and \textit{Feedback Backpropagation} represent the forward and backward information flow, respectively. 
In the former phase, the model reveals the effect of the selected token on the final generation through deduction and evaluation. 
The model's self-evaluation feedback then, in turn, improves the token selection at the chaotic point. 
Essentially, self-evaluation decoding enhances the model's generation through its evaluation.
Perceptibly, accurate feedback is quite important for self-evaluative decoding to enhance generation. 
Recent studies~\citep{23-autoj,24-chatglm-math} have shown LLMs can be trained to build good evaluation ability. 
Here we propose a simple but effective data augmentation approach to efficiently construct instruction finetuning data to empower the model's evaluation ability.


\textbf{Multi-model generation ensembling}. 
Given a downstream task dataset $\mathcal{D}=\{(q_i,a_i)\}_{i=1}^{n}$ and a collection of LLMs $\mathcal{M}=\{M_i\}_{i=1}^{m}$, we use each model in $\mathcal{M}$ to answer each question in $\mathcal{D}$, which in turn yields model responses $\{r_{eval}^{i}\}_{i=1}^{n*m}$ in a CoT format~\citep{22-cot}. 
We compare these generated responses with the question answers to obtain the final augmented dataset $\mathcal{D}_{eval}=\{(q_{eval}^{l},r_{eval}^{l},j_{eval}^{l})\}_{l=1}^{n*m}$, where the judgment $j_{eval}^{l}$ denotes the correctness of the response $r_{eval}^{l}$.
In this way, each $q_i$ in $\mathcal{D}$ receives multiple, CoT-formatted responses from different LLMs with different correctnesses.
Furthermore, we extract all the correct responses in $\mathcal{D}_{eval}$ thereby constructing the dataset $\mathcal{D}_{gen}=\{(q_{gen}^l,r_{gen}^l)\}\}_{l=1}^{\# \text{correct}}$. 
$\mathcal{D}_{eval}$ and $\mathcal{D}_{gen}$ are used to train the model's evaluation and generation abilities on the downstream tasks, respectively.

\textbf{Training}.
Similar to previous studies, we use $\mathcal{D}_{eval}$ to train the model's evaluation ability. 
Specifically, the model learns to output $j_{eval}^{l}$ given $q_{eval}^l$ and $r_{eval}^{l}$:
\[
\mathcal{L}_{eval}^l(\Theta) = -\log\mathbb{P}_\Theta\left(j_{eval}^l|q_{eval}^l,r_{eval}^l\right)
\]
$\Theta$ represents the model's parameters. 
The model's generation ability is developed by the loss function:
\[
\mathcal{L}_{gen}^l(\Theta) = -\log\mathbb{P}_\Theta\left(r_{gen}^l|q_{gen}^l\right)
\]
Here, we adopt multi-task learning to train the model to build the generation and evaluation abilities simultaneously:
\[
\min_\Theta \mathcal{L}(\Theta) =
\min_\Theta 
\sum_{i=1} \mathcal{L}_{eval}^{i}(\Theta) + 
\sum_{j=1} \mathcal{L}_{gen}^{j}(\Theta)
\]
Data sampled from $\mathcal{D}_{eval}$ and $\mathcal{D}_{gen}$ respectively and the ratio is maintained around 1:1. 
They are mixed to build the final training set.
Meanwhile, the sampled data from $\mathcal{D}_{eval}$ has approximately the same number of samples with $j_{eval}=1$ and $j_{eval}=0$. More details about $j_{eval}$ are available in Appendix~\ref{sec:appendix-judgement-inference}.


We build the dataset described above and train the model in a multi-task learning manner.
This method enhances the model's abilities in two ways. 
By integrating multiple model generations, representing diverse thoughts on the same question, the model can learn the nuanced differences and thus know how to analyze the question from several perspectives. 
Meanwhile, recent studies~\citep{23-metamath, 23-learning-from-mistakes} have shown that understanding a question from the opposite is an effective way to enhance the model's comprehension of the downstream tasks.
Therefore, learning to judge the correctness of an answer from two sides should also be an effective way to boost the model's abilities.

\section{Experiment}
\label{sec:experiment}
\begin{table*}[!htb]
    \captionsetup{aboveskip=8pt, belowskip=0pt} 
    \caption{The performances of different decoding strategies for downstream tasks. 
    Models with marker `*' are trained only on $\mathcal{D}_{gen}$, while those without marker `*' are trained on $\mathcal{D}_{gen}$ and $\mathcal{D}_{aug}$. 
    We keep the training steps of them equal for comparison.
    Experiments of sampling-based methods are repeated 5 times to derive the \texttt{mean} and \texttt{std}.
    \textbf{Bold} and \uline{underline} denote the best and second best.
    }
    \label{tab:main-result}
    \renewcommand\arraystretch{1.2}
    \centering
    \scalebox{0.9}{
        \begin{tabular}{c|c|ccc}
            \Xhline{1pt} 
            \multirow{2}{*}{\textbf{LLMs}} &  \multirow{2}{*}{\textbf{Decoding Strategies}} & \multicolumn{3}{c}{\textbf{Acc.}} \\ 
            & 
            & 
            \textbf{XieZhi} & 
            \textbf{HotpotQA} &  
            \textbf{GSM8K} \\ 
            \hline
            \multirow{2}{*}{\rotatebox{0}{\texttt{falcon-7b-instruct}}*} 
            & \text{greedy search} & 0.251 & 0.725 & 0.167 \\ 
            & \text{beam search} & 0.265 & 0.722 & 0.175 \\ 
            \hline
            \multirow{5}{*}{\rotatebox{0}{\texttt{falcon-7b-instruct}}} 
            & \text{greedy search} & 0.272 & 0.735 & 0.225 \\ 
            & \text{beam search} & 0.292 & 0.736 & 0.216 \\ 
            & \text{nuclear sampling} & 0.248$\pm$0.019 & 0.706$\pm$0.014 & 0.204$\pm$0.022 \\ 
            & \text{SED, ratio-based} & \textbf{0.338} & \textbf{0.750} & \textbf{0.256} \\ 
            & \text{SED, entropy-based} & \uline{0.326} & \uline{0.745} & \uline{0.244} \\ 
    
            \hline
            \multirow{2}{*}{\rotatebox{0}{\texttt{llama2-7b-chat-hf}}*} 
            & \text{greedy search} & 0.304 & 0.794 & 0.326 \\ 
            & \text{beam search} & 0.329 & 0.815 & 0.352 \\ 
            \hline
            \multirow{6}{*}{\rotatebox{0}{\texttt{llama2-7b-chat-hf}}}
            & \text{greedy search} & 0.312 & 0.804 & 0.372 \\ 
            & \text{beam search} & 0.336 & 0.808 & 0.396 \\ 
            & \text{nuclear sampling} & 0.255$\pm$0.022 & 0.783$\pm$0.017 &  0.342$\pm$0.025 \\ 
            & \text{DoLa} & 0.286 & 0.725 & 0.323 \\ 
            & \text{SED, ratio-based} & \textbf{0.415} & \textbf{0.822} & \textbf{0.421} \\ 
            & \text{SED, entropy-based} & \uline{0.395} & \uline{0.820} & \uline{0.405} \\ 
    
            \hline
            \multirow{2}{*}{\rotatebox{0}{\texttt{gemma-7b-it}}* }
            & \text{greedy search} & 0.294 & 0.808 & 0.462 \\ 
            & \text{beam search} & 0.288 & 0.801 & 0.470 \\ 
            \hline
            \multirow{5}{*}{\rotatebox{0}{\texttt{gemma-7b-it}} } 
            & \text{greedy search} & 0.323 & 0.840 & 0.472 \\ 
            & \text{beam search} & 0.330 & 0.848 & 0.480 \\ 
            & \text{nuclear sampling} & 0.288$\pm$0.018 & 0.822$\pm$0.017 & 0.456$\pm$0.018 \\ 
            & \text{SED, ratio-based} & \uline{0.330} & \textbf{0.865} & \textbf{0.510} \\ 
            & \text{SED, entropy-based} & \textbf{0.335} & \uline{0.862} & \uline{0.498} \\ 
            \Xhline{1pt}
        \end{tabular}
    }
\end{table*}


\subsection{Datasets}
\label{sec:datasets}
We use three datasets: (i) the HotpotQA subset~\citep{18-hotpotqa}, processed into the SQuAD format, from the MRQA 2019 Shared Task~\citep{19-mrqa}. Retrieval is not needed, allowing LLMs to focus on answering questions in light of the references; 
(ii) XieZhi~\citep{24-xiezhi}, a comprehensive evaluation suite with a multiple-choice format, aiming to test diverse subject knowledge of LLMs; 
and (iii) GSM8K~\citep{21-gsm8k}, a popular mathematical reasoning benchmark for evaluating LLMs' mathematical abilities, consists of high-quality elementary school math word problems. 
They cover diverse application scenarios of LLMs, such as multi-hop reasoning, multi-disciplinary knowledge comprehension, and mathematical reasoning.


\subsection{Baselines and metrics}
\label{sec:baselines-metrics}
We choose greedy search, beam search, and nuclear sampling as our baselines.
They are widely used decoding methods in LLM inference.
We also compare with DoLa~\citep{23-dola}, a contrastive decoding that alleviates hallucinations. 
\texttt{falcon-7b-instruct}~\citep{23-falcon}, \texttt{llama2-7b-chat-hf}~\citep{23-llama2}, and \texttt{gemma-7b-it}~\citep{23-gemini} are trained on all datasets through supervised fine-tuning described in Section~\ref{sec:eval-abi-cons}.
We evaluate the performance of our proposed self-evaluation decoding by comparing it with other decoding baselines across different fine-tuned models. 
Due to the official open-source code of DoLa currently only supporting the Llama series models, we use it as a baseline only on Llama2.

We compare the answers extracted from the model responses with the annotations to calculate accuracy, thereby measuring the performance of different decoding methods. 
We uniformly sample 1,000 instances for testing from each dataset to manage experimental costs.

\subsection{Experimental settings}
\label{sec:implementation-details}
As mentioned in Section~\ref{sec:eval-abi-cons}, we employ multiple LLMs to augment downstream datasets, including \texttt{baichuan2-7b-chat}~\citep{23-baichuan2}, \texttt{qwen-7b-chat}~\citep{23-qwen}, \texttt{mistral-7b-instruct-v0.2}~\citep{23-mistral} and \texttt{metamath-mistral-7b}~\citep{23-metamath} (specifically for GSM8K). 
Prompt details can be found in Appendix~\ref{sec:appendix-prompt-template}.
For every question in each dataset, we sample multiple responses from each model and filter out duplicate and incorrectly formatted responses. 
We extract the answers through regular expressions from those retained responses and compare them with annotations to determine their correctness.
Finally, according to the operations described in Section~\ref{sec:eval-abi-cons}, $\mathcal{D}_{eval}$ and $\mathcal{D}_{gen}$ are obtained.
We train the models on 2 A800 GPUs using ZeRO~\citep{20-zero} stage 3 from DeepSpeed~\citep{20-deepspeed}. 
We adopt AdamW~\citep{14-adamw} as the optimizer and set the batch size to 8 with a maximum sequence length of 1,024 and one training epoch. 
The number of beams $n_b$ of beam search and the temperature $\tau$ and $p$ of nucleus sampling are set to $n_b=3$ and $\tau=0.7,p=0.9$, respectively.
More hyperparameters for the experiment are reported in Appendix~\ref{sec:appendix-hyper-parameter-selection}.

\subsection{Main results}
\label{sec:main-results}
Table~\ref{tab:main-result} presents the three LLMs' results that employ various decoding methods.
Beam search and nuclear sampling explore a wider solution space than greedy search, and nuclear sampling increases diversity.
However, compared to greedy search, nuclear sampling performs poorly with large fluctuations, indicating nuclear sampling is not a good choice in QA scenarios with standard answers. 
On the other hand, beam search is superior to greedy search in most cases, suggesting that a larger solution space helps to get a more accurate response.
Alleviating hallucination seems not helpful, since DoLa performs poorly consistently.
Both entropy-based and ratio-based SED outperform other baselines across multiple models and datasets, 
especially for \texttt{falcon} and \texttt{llama2} on XieZhi and GSM8K.
As shown, in most cases, the ratio-based SED performs slightly better than the entropy-based. 
We attribute this to the former's more specific constraints on token probabilities, which help avoid misjudging chaotic points. 
Additionally, adjusting the ratio threshold is much more convenient than the entropy threshold. 
Thus, ratio-based SED is more practical compared to entropy-based.
In subsequent experiments, ratio-based SED is the default implementation.
We observe that the improvement in model performance for SED, relative to the baseline models, varies across different datasets. 
Specifically, the performance increment is more pronounced in the XieZhi and GSM8K  compared to HotpotQA. 
Chaotic points in GSM8K often occur at numbers and operators, whereas in XieZhi, they frequently appear at the options, as shown in Table~\ref{tab:case-study-gsm8k} and Table~\ref{tab:case-study-xiezhi}.
The impact of these chaotic points on the accuracy of model responses is significant. 
SED allows the model to effectively select tokens at these positions, thereby avoiding incorrect answers.
It is important to note that when $\mathcal{D}_{eval}$ is incorporated into the training process, models' accuracies increase to varying extents. 
This suggests that enhancing a model's evaluation ability could indirectly improve its generation ability, which may be related to the introduction of incorrect responses in $\mathcal{D}_{eval}$ to enable the model to learn how to identify flaws and thereby enhance its ability from the opposite perspective~\citep{23-learning-from-mistakes}. 
More cases that demonstrate SED's effect are available in Appendix~\ref{sec:appendix-case-study}.

\subsection{Analysis of evaluation quality}
\label{sec:analysis-feedback-quality}
\begin{table}
        \captionsetup{aboveskip=8pt, belowskip=-20pt}
        \small\caption{The effects of feedback quality on generation.
        We use trained \texttt{llama2-7b-chat-hf} for generation and speculation, and other stronger LLMs for evaluation.
        Set $\alpha=1$ to fully test the evaluation feedback quality.
        `\textbf{Direct gen.}' indicates the accuracies the LLMs achieve through directly answering questions with greedy search.
        `\textbf{SE. gen.}' means the accuracies that \texttt{llama2-7b-chat-hf} achieves through self-evaluation decoding with various LLMs as the feedback providers respectively.
        }
        \label{tab:feedback-quality}
        \renewcommand\arraystretch{1.1}
        \centering
        \scalebox{1}{
        \begin{tabular}{c|cc}
            \Xhline{1pt}
             \multirow{2}{*}{\textbf{LLMs (Feedback Source)}} & \multicolumn{2}{c}{\textbf{XieZhi Acc.}} \\
            & \textbf{Direct gen.} & \textbf{SE. gen.} \\
            \hline
            \texttt{llama2-7b-chat-hf} (self) & 0.312 & 0.380\\
            \hline
            \texttt{llama3-8b-instruct} & 0.644 & 0.505\\
            \texttt{llama3-70b-instruct} & 0.663 & 0.584 \\
            \texttt{yi-34b-chat} & 0.665 & 0.620 \\
            \texttt{qwen1.5-72b-chat} & 0.681 & 0.588  \\
            \Xhline{1pt}
        \end{tabular}
    }
\end{table}

Intuitively, the evaluation quality heavily influences the model's final response. 
Ideally, if the evaluation quality is sufficiently high, the model can generate an optimal response by completely following the evaluation feedback whenever it encounters a chaotic point. 
To investigate the effect of evaluation quality, we adopt other stronger LLMs as the speculation evaluators to generate more accurate evaluation feedback. 
The results are shown in Table~\ref{tab:feedback-quality}.
In addition to llama2's self-evaluation, we also use \texttt{llama3-8b/70b-instruct}~\citep{24-llama3}, \texttt{yi-34b-chat}~\citep{24-yi}, and \texttt{qwen1.5-72b-chat}~\citep{23-qwen} to evaluate the speculation. 
We choose XieZhi~\citep{24-xiezhi} for testing.
As mentioned before, the option token in XieZhi is an important chaotic point.
The evaluation feedback at the option token can significantly affect the decoding result. 
Here, we use these models to directly answer questions and then obtain their accuracies (\textbf{Direct gen.} in Table~\ref{tab:feedback-quality}) to approximate their evaluation quality, as these accuracies are the most intuitive representation of LLMs' capabilities. 
Results in Table~\ref{tab:feedback-quality} (\textbf{SE. gen.}) clearly show that llama2's accuracy is significantly improved when we use a more powerful model to provide more accurate evaluation feedback. 
This demonstrates evaluation quality is crucial in SED. 
The more accurate the evaluation feedback, the better SED works.
The results also suggest that we could make use of tools to further enhance the evaluation quality and thus further improve the decoding in other tool-friendly application scenarios.

\subsection{Analysis of generation-speculation alignment}
\label{sec:analysis-speculation-alignment}
\begin{table}
    \captionsetup{aboveskip=8pt, belowskip=-5pt}
    \small\caption{The effects of alignment and misalignment between generation and speculation. 
    \texttt{llama2-7b-chat-hf} for generation, speculation, and evaluation with $\alpha=1$. 
    The results with \textbf{Speculation} being `-' indicate they are directly obtained through baseline decoding methods without applying our proposed self-evaluation framework.
    }
    \label{tab:decoding-consistency}
    \renewcommand\arraystretch{1.1}
    \centering
    \scalebox{1}{
        \begin{tabular}{c|c|cc}
            \Xhline{1pt} 
            \multirow{2}{*}{\textbf{Generation}} &  \multirow{2}{*}{\textbf{Speculation}} & \multicolumn{2}{c}{\textbf{Acc.}} \\ 
            & 
            & 
            \textbf{HotpotQA} &  
            \textbf{GSM8K} \\ 
            \hline
            \multirow{3}{*}{\rotatebox{0}{greedy search}} 
            & -  & 0.804 & 0.372  \\ 
            & \text{greedy search} & \textbf{0.816} & \textbf{0.416}  \\ 
            & \text{beam search} & 0.802 & 0.378  \\ 
            \hline
            \multirow{3}{*}{\rotatebox{0}{beam search}}
            & - & 0.808 & 0.396  \\ 
            & \text{greedy search} & 0.805 & 0.388  \\ 
            & \text{beam search} & \textbf{0.819} & \textbf{0.412} \\  
            \hline
            \multirow{3}{*}{\rotatebox{0}{nuclear sampling}}
            & - & 0.783 & 0.342 \\ 
            & \text{greedy search} & 0.778 & 0.337 \\ 
            & \text{beam search} & 0.791 & 0.344 \\ 
            \Xhline{1pt}
        \end{tabular}
    }
        \vspace{-0.2cm}
\end{table}

Besides \textit{evaluation}, \textit{speculation} is also an important step for SED. 
As described in Section~\ref{sec:speculation-and-evaluation}, \textit{speculation} refers to inferring the text that the model will subsequently generate after token $t_i$ is selected. 
This requires that the speculation needs to be accurate enough. 
More specifically, the model needs to have the same token selected at speculation as at generation, i.e., the decoding methods at both stages should be aligned. 
To investigate the impact of the decoding methods' alignment, we check various decoding combinations in the generation and speculation phases. 
As shown in Table~\ref{tab:decoding-consistency}, the model performance is improved when the decoding methods at generation and speculation are aligned. 
Otherwise, SED provides a limited gain to the model and sometimes negatively affects the performance.
This implies that SED works when the speculation and generation are aligned. 
The misalignment could give the model a misleading, which in turn affects the model's final response.
We choose the `greedy + greedy' combination as the implementation for SED in Table~\ref{tab:main-result} since it requires less computation than `beam + beam' with similar performance.


\vspace{-0.3em} 
\section{Conclusion}
\label{sec:conclusion}
\vspace{-0.3em} 
In this paper, we propose a decoding method called self-evaluation decoding, SED, to optimize LLMs' token selection at chaotic points during decoding so that the most appropriate token can be selected.
SED first detects the chaotic points and then obtains the self-evaluation probability of each token at the chaotic points through two steps: \textit{speculation} and \textit{evaluation}, and finally combines the self-evaluation probabilities with the token probabilities to determine the optimal token.
Finally, experimental results and case studies across various tasks demonstrate the effectiveness of our method.

\vspace{-0.3em} 
\section{Limitation}
\label{sec:limitation}
\vspace{-0.3em}
Our method outperforms other baselines across different models on various tasks, illustrating its effectiveness. 
However, there remain some limitations. 
First, judgment $j_{eval}^l$ in $\mathcal{D}_{eval}$ is relatively simple now, lacking of CoT or other enhanced information.
Second, the introduction of \textit{speculation} makes the decoding slower since more computations are required for selecting the optimal token. 
Indeed, it is a trade-off between inference efficiency and quality.
\cite{17-thinking-fast-and-slow} mentions two types of decision-making behaviors: \textit{fast thinking} and \textit{slow thinking}. 
\textit{Slow thinking} is responsible for solving complex problems that are hard for \textit{fast thinking}, needing additional computational resources for logical and rational thinking.
In other words, to optimize LLMs' token selection at chaotic points, we ought to allow LLMs to think slowly, and exchange the extra inference time for better generation, which is exactly the core contribution of this paper.
Regarding the additional time cost introduced by \textit{slow thinking}, we could reduce it through methods such as reusing computation.
We leave the optimizations for evaluation ability enhancement and decoding speed as our future work.

\bibliographystyle{plainnat}
\bibliography{neurips_2024}
\newpage







\appendix



\section{Judgement representation and inference}
\label{sec:appendix-judgement-inference}
$j_{eval}^l$ in Section~\ref{sec:eval-abi-cons} is represented in numeric.
Nevertheless, we need to convert it to text for training.
Denote $t_{j_{eval}^l}$ as the actual output that the model learns to generate given the context of $q_{eval}^l$ and $r_{eval}^l$. When we maximize $\mathbb{P}_\Theta\left(j_{eval}^l|q_{eval}^l,r_{eval}^l\right)$, we are actually maximizing the probability
\[
\mathbb{P}_\Theta\left(t_{j_{eval}^l}|q_{eval}^l,r_{eval}^l\right), \quad t_{j_{eval}^l} = \left\{\begin{matrix}
\text{``[[Y]], the answer is correct.''} & j_{eval}^l=1 \\
\text{``[[N]], the answer is incorrect.''} & j_{eval}^l=0 \\
\end{matrix}\right.
\]
Therefore, in terms of calculating $p_{se}^{(j)}$ in Section~\ref{sec:speculation-and-evaluation}, we are actually discussing about the probability
\[
p_{se}^{(j)} := 
\mathbb{P}\left(J_{se}^{(j)}=1|Q, R_{se}^{(j)}\right) = 
\mathbb{P}\left(\text{``[[}\cdot\text{]]''}=\text{``[[Y]]''}|Q, R_{se}^{(j)}\right)
\]
More specifically, we take the model's output \texttt{logits} at the position behind token ``[['' and calculate the probability of token `Y' being selected, which is essentially the $p_{se}^{(j)}$ we want.

\section{Prompt templates}
\label{sec:appendix-prompt-template}
This section shows the prompt templates for generating multi-model responses, model inference, and self-evaluation.
The few-shot templates for XieZhi and GSM8K are shown in Table~\ref{tab:prompt_for_xiezhi} and Table~\ref{tab:prompt_for_gsm8k}, while the zero-shot template for HotpotQA is shown in Table~\ref{tab:prompt_for_hotpotqa}.
We apply three prompt templates shown in Table~\ref{tab:prompt_for_model_inference_and_evaluation} to model inference on different datasets.
The self-evaluation prompt for \textit{Evaluation} step in self-evaluation decoding is also shown in Table~\ref{tab:prompt_for_model_inference_and_evaluation}.

\section{Case study on the behavior of self-evaluative decoding}
\label{sec:appendix-case-study}
To demonstrate how SED can guide LLMs to obtain more accurate responses and thus improve performances, we use the trained \texttt{llama2-7b-chat-hf} through both greedy search and self-evaluation decoding, SED, to present three generated cases in Table~\ref{tab:case-study-xiezhi},~\ref{tab:case-study-hotpotqa},~and~\ref{tab:case-study-gsm8k}.

The texts generated by the model using SED and greedy search have the same prefix, and we label them with underlines. 
In addition, we use light red and light green to label the tokens selected by greedy search and SED at the first chaotic point respectively.
Other colors annotate the key information the model can pay attention to in the question or references.
For the XieZhi problem in Table~\ref{tab:case-study-xiezhi}, the first chaotic point appears on the option token.
Different options could directly lead to the model outputting completely different responses. 
When the model applies greedy search for decoding, it mistakenly selected option \texttt{D} and subsequently generates incorrect explanations.
While SED makes the model carefully reflect on each option based on speculation and evaluation, making the model choose the correct option \texttt{B} since it has the highest propensity score.
For the HotpotQA problem in Table~\ref{tab:case-study-hotpotqa}, both greedy search and SED enable the model to correctly analyze the person described in the problem.
At the first chaotic point, the greedy search selects token `\textbackslash n' and directly takes the person's name as the answer, while SED guides the model to answer the person's distinction, which is exactly the correct answer, by restating the question.
For the GSM8K question in Table~\ref{tab:case-study-gsm8k}, the greedy search selects the number `4' at the first chaotic point and mistakenly treats the period ``\texttt{3 hours}'' as a distance, leading to an incorrect answer.
SED selects the number `1' to guide the model in calculating the distance traveled by Tom thereby generating the correct answer in the end. 
These three real cases vividly demonstrate the effectiveness of Self-Evaluation Decoding.

\section{Experimental details}
\label{sec:appendix-hyper-parameter-selection}
There are quite a few possible combinations of hyper-parameters.
It is unfeasible to search for the best among them through extensive experiments.
Therefore, we pre-determine some through small-scale pilot experiments on a sampled subset for each dataset.
We empirically determine $\delta_r$ to be $0.8$ and $\delta_e$ to be $2.25$.
Branching factor $b$ and top $k$ tokens are another two parameters we need to consider.
Empirically, we check several model responses given several questions sampled from the datasets and find many detected chaotic points among the responses.
We assume that the chaotic points at the beginning of the responses are more important than those at the end since the former can affect more tokens than the latter.
On the other hand, the probability mass for token selection is mainly concentrated on top-5 tokens in many cases.
Considering the balance between cost and effectiveness, we set $b=5$ and $k=5$.
As for the fusion coefficient $\alpha$, we run a grid search with a subset of the data (30 questions per dataset), using \texttt{llama2-7b-chat-hf}.
We experiment with $\alpha = [0.1, 0.2, \dots, 1]$ and find that $\alpha=0.8$ is best for XieZhi and GSM8K while $\alpha=0.6$ is best for HotpotQA, indicating the model's evaluation ability varies on different datasets.



\begin{table*}[!h]
  \caption{
        A case from XieZhi shows that self-evaluation decoding can help the model reflect on the chosen option.
      }
    \label{tab:case-study-xiezhi}
    \vspace{0.1cm}
  \centering
  \small
  \begin{tabular}{p{0.15\linewidth} p{0.85\linewidth}}
    \toprule
    \textbf{Question}  & What is \colorbox{lightYellow}{not explicitly mentioned} in the "Three People's Principles" is \_\_\_\_ \\
    & A: Against the feudal rule of the Qing Dynasty. \\
    & B: Oppose foreign imperialism.\\
    & C: Establishment of the Republic of China. \\
    & D: Equal distribution of land. \\
    \midrule
    \textbf{Greedy Search}  & \uline{The answer should be} \colorbox{lightRed}{\texttt{D}}, which is \texttt{Equal distribution of land}. While the principle of people's livelihood aimed to ensure economic and social welfare for the people, it did not specifically mention the equal distribution of land. \textcolor{red}{\usym{2718}}\\
    \midrule
    \textbf{Self-Evaluation Decoding}  & \uline{The answer should be} \colorbox{lightGreen}{\texttt{B}}, which is \texttt{Oppose foreign imperialism}. While the principles primarily focused on nationalism, democracy, and people's livelihood, the explicit opposition to foreign imperialism was not a central theme within these principles. \textcolor{green}{\usym{2714}}\\
    \midrule
    \textbf{Answer}  & \texttt{B: Oppose foreign imperialism}.\\
    \bottomrule
  \end{tabular}
\end{table*}
\clearpage
\begin{table*}[!h]
  \caption{
        A case from HotpotQA demonstrates that self-evaluative decoding can help the model avoid generating irrelevant answers.
      }
    \label{tab:case-study-hotpotqa}
    \vspace{0.1cm}
  \centering
  \small
  \begin{tabular}{p{0.15\linewidth} p{0.85\linewidth}}
    \toprule
     \textbf{Question} & What \colorbox{lightYellow}{distinction} is held by the former NBA player who was a member of the Charlotte Hornets during their 1992-93 season and was head coach for the WNBA team Charlotte Sting?\\
    \midrule
    \textbf{Reference 1}  & 1992–93 Charlotte Hornets season \\
    & The 1992–93 NBA season was the fifth season for the Charlotte Hornets in the National Basketball Association. In the 1992 NBA draft, the Hornets selected center Alonzo Mourning out of Georgetown with the second overall pick.  With the addition of Mourning, along with second-year star \colorbox{lightYellow}{Larry Johnson and Muggsy Bogues}, the Hornets struggled around .500 for most of the season, but won 9 of their final 12 games finishing their season third in the Central Division with a 44–38 record, and qualified for their first ever playoff appearance.  The team became the first of the four expansion franchises of the late 1980s to win a playoff series, when Mourning sunk a 20 ft jumper at the buzzer in Game 4 of the first round playoff series against the 4th-seeded Boston Celtics.  However, in the semifinals, the Hornets were eliminated by the New York Knicks in five games. \\
    \textbf{Reference 2} & \colorbox{lightYellow}{Muggsy Bogues} \\
    & Tyrone Curtis Muggsy Bogues (born January 9, 1965) is an American retired basketball player.  \colorbox{lightYellow}{The shortest player ever to play in the National Basketball Association}, the 5 ft Bogues played point guard for four teams during his 14-season career in the NBA.  He is best known for his ten seasons with the \colorbox{lightYellow}{Charlotte Hornets} and also played for the Washington Bullets, Golden State Warriors, and Toronto Raptors. After his NBA career, he served as \colorbox{lightYellow}{head} \colorbox{lightYellow}{coach} of the now-defunct WNBA team Charlotte Sting. \\
    \midrule
    \textbf{Greedy Search}  & \uline{The former NBA player who was a member of the Charlotte Hornets during their 1992-93 season and was head coach for the WNBA team Charlotte Sting is Muggsy Bogues.}\colorbox{lightRed}{\textbackslash n}Therefore, the answer is: \texttt{Muggsy Bogues}. \textcolor{red}{\usym{2718}}\\
    \midrule
    \textbf{Self-Evaluation Decoding}  & \uline{The former NBA player who was a member of the Charlotte Hornets during their 1992-93 season and was head coach for the WNBA team Charlotte Sting is Muggsy Bogues.} \colorbox{lightGreen}{ The} question asks for the distinction that he holds. According to Reference 2, Muggsy Bogues is the shortest player ever to play in the National Basketball Association. This is the distinction that he holds.\\
    & Therefore, the answer is: \texttt{Muggsy Bogues is the shortest player ever to play in the National Basketball Association}. \textcolor{green}{\usym{2714}}\\
    \midrule
    \textbf{Answer}  & \texttt{the shortest player ever to play in the National Basketball Association}.\\
    \bottomrule
  \end{tabular}
\end{table*}
\clearpage


\begin{table*}[!h]
  \caption{
        A case from GSM8K shows that self-evaluation decoding can help the model check the unsure numbers.
      }
    \label{tab:case-study-gsm8k}
    \vspace{0.1cm}
  \centering
  \small
  \begin{tabular}{p{0.15\linewidth} p{0.85\linewidth}}
    \toprule
    \textbf{Question}  & Tom's ship can \colorbox{lightYellow}{travel at 10 miles per hour}. He is \colorbox{lightYellow}{sailing from 1 to 4 PM}. \\
    & He then \colorbox{lightYellow}{travels back at a rate of 6 mph}. \\
    & How long does it take him to get back? \\
    \midrule
    \textbf{Greedy Search}  &  \uline{Step1: Tom's ship travels at 10 miles per hour from 1 to 4 PM. So, the distance he travels is }\colorbox{lightRed}{4} - 1 = 3 hours.\\
    & Step2: Then, he travels back at a rate of 6 mph. So, the time it takes him to get back is 3 hours / 6 mph = 0.5 hours. \\
    & Step3: Finally, we add the time it takes him to travel forward and back to get the total time it takes him to complete the journey. \\
    & So, the total time is 3 hours + 0.5 hours = 3.5 hours. \\
    & So, the final answer is \texttt{3.5}. \textcolor{red}{\usym{2718}}\\
    \midrule
    \textbf{Self-Evaluation Decoding}  &  \uline{Step1: Tom's ship travels at 10 miles per hour from 1 to 4 PM. So, the distance traveled is }\colorbox{lightGreen}{1}0 miles/hour * 3 hours = 30 miles.\\ 
    & Step2: Then, Tom travels back at a rate of 6 mph. So, the time taken to travel back is 30 miles / 6 mph = 5 hours. \\
    & So, the final answer is \texttt{5}. \textcolor{green}{\usym{2714}}\\
    \midrule
    \textbf{Answer}  & He was traveling at full speed for 4-1=<<4-1=3>>3 hours\\
    & So he went 3*10=<<3*10=30>>30 miles\\
    & That means it takes 30/6=<<30/6=5>>5 hours to get back\\
    & \#\#\#\# \texttt{5}\\
    \bottomrule
  \end{tabular}
\end{table*}
\clearpage

\begin{table*}[h]
    \caption{Few-shot prompt used for XieZhi}
    \label{tab:prompt_for_xiezhi}
    \vspace{-0.2cm}
    \centering
    \resizebox{1\textwidth}{!}
    {
    \begin{tabular}{p{1\textwidth}}\\
    \toprule
    \begin{minipage}{\linewidth}
        \setstretch{1} 
         \textcolor{gray}{/* }\textcolor{gray}{\textit{Prompt Template }}\textcolor{gray}{*/}\\
        Select the answer to the target question from the options based on the knowledge you have learned.\\
        \\
        There are some examples that you can refer to: \\
        --- \\
        \{\texttt{demonstrations}\} \\
        --- \\
        \\
        You should answer the following question in a similar way to the examples above:\\
        Question: \{\texttt{question\_and\_options}\} \\
        Answer: \\
        \\
        \textcolor{gray}{/* }\textcolor{gray}{\textit{Example for } \texttt{demonstrations}}\textcolor{gray}{*/}\\
        \texttt{demonstrations} = \textasciigrave\textasciigrave\textasciigrave \\
        Question: China was the first country in the world to grow rice, as you can see from \_\_\_\_.\\
        Options: A: Lantian Man ruins, B: Site of Dawenkou, C: Banpo site, D: Hemudu Ruins  \\
        Answer: The answer should be [[D]], which is Hemudu Ruins. Archaeological excavations at this site have uncovered tools, pottery, and rice grains, indicating that rice was indeed cultivated there. \\
        ---\\
        The relocation of a capital is a major event in a country, and the one who moved the capital to accept the advanced culture is \_\_\_\_\\
        Options: A: Pan Geng moved Yin, B: Emperor Xiaowen of Northern Wei moved his capital to Luoyang, C: King Ping moved the capital to Luoyi, D: The capital of Soviet Russia was moved to Moscow \\
        Answer: The answer should be option [[B]], which is Emperor Xiaowen of Northern Wei moved his capital to Luoyang. Luoyang was a center of culture and civilization during that time, and by moving the capital there, Emperor Xiaowen aimed to strengthen the Northern Wei dynasty by embracing the cultural, economic, and political advancements of the region. \\
        ---\\
        (more few-shot demonstrations) \dots \\
        \textasciigrave\textasciigrave\textasciigrave
    \end{minipage}\\
    \bottomrule
    \end{tabular}
    }
\end{table*}
\clearpage
\begin{table*}[h]
    \caption{Zero-shot prompt used for HotpotQA}
    \label{tab:prompt_for_hotpotqa}
    \vspace{-0.2cm}
    \centering
    \resizebox{1\textwidth}{!}
    {
    \begin{tabular}{p{1\textwidth}}\\
    \toprule
    \begin{minipage}{\linewidth}
        \setstretch{1} 
         \textcolor{gray}{/* }\textcolor{gray}{\textit{Prompt Template }}\textcolor{gray}{*/}\\
        You are professional at reading comprehension. \\
        \\
        The following is one or several reference documents for you to comprehend and memory: \\
        --- \\
        \{\texttt{reference\_docs}\} \\
        --- \\
        \\
        The answer to the following question can be inferred through the above reference(s).\\
        Question: \{\texttt{question}\} \\
        \\
        Provide the answer to the given question through step-by-step reasoning. After your reasoning, output your final answer in the format of [[YOUR ANSWER]].\\
        Make sure the final answer is enclosed by the double square brackets.\\
        \\
        Let's think step by step.
    \end{minipage}\\
    \bottomrule
    \end{tabular}
    }
\end{table*}
\clearpage
\begin{table*}[htbp]
    \caption{Few-shot prompt used for GSM8K}
    \label{tab:prompt_for_gsm8k}
    \vspace{-0.2cm}
    \centering
    \resizebox{1\textwidth}{!}
    {
    \begin{tabular}{p{1\textwidth}}\\
    \toprule
    \begin{minipage}{\linewidth}
        \setstretch{1} 
         \textcolor{gray}{/* }\textcolor{gray}{\textit{Prompt Template }}\textcolor{gray}{*/}\\
        Please help me solve the math problem step by step. You must give your final answer at the end of response like `So, the final answer is \$ your answer'. I will provide examples for you:\\
        \\
        --- \\
        \{\texttt{demonstrations}\} \\
        --- \\
        \\
        Next, please answer this math problem: \\
        Question: \{\texttt{question}\} \\
        Answer: \\
        \\
        \textcolor{gray}{/* }\textcolor{gray}{\textit{Example for } \texttt{demonstrations}}\textcolor{gray}{*/}\\
        \texttt{demonstrations} = \textasciigrave\textasciigrave\textasciigrave \\
        Question: Ken created a care package to send to his brother, who was away at boarding school.  Ken placed a box on a scale, and then he poured into the box enough jelly beans to bring the weight to 2 pounds.  Then, he added enough brownies to cause the weight to triple.  Next, he added another 2 pounds of jelly beans.  And finally, he added enough gummy worms to double the weight once again.  What was the final weight of the box of goodies, in pounds? \\
        Answer: Step1: To the initial 2 pounds of jelly beans, he added enough brownies to cause the weight to triple, bringing the weight to 2*3=6 pounds.\\
        Step2: Next, he added another 2 pounds of jelly beans, bringing the weight to 6+2=8 pounds.\\
        Step3: And finally, he added enough gummy worms to double the weight once again, to a final weight of 8*2=16 pounds.\\
        So, the final answer is \$16.\\
        ---\\
        (more few-shot demonstrations) \dots\\
        \textasciigrave\textasciigrave\textasciigrave
    \end{minipage}\\
    \bottomrule
    \end{tabular}
    }
\end{table*}
\clearpage

\begin{table*}[!h]
  \caption{
        Prompts for model inference and self-evaluation.
      }
    \label{tab:prompt_for_model_inference_and_evaluation}
    \vspace{0.1cm}
  \centering
  \small
  \begin{tabular}{p{1\linewidth}}
    \toprule
    \textcolor{gray}{/* }\textcolor{gray}{\textit{ Model inference prompt for XieZhi }}\textcolor{gray}{*/} \\
     Select the answer to the question from the options based on the knowledge you have learned. \\
     Question: \{question\} \\
     Options: \{options\} \\
     Answer: \\
    \midrule
    \textcolor{gray}{/* }\textcolor{gray}{\textit{ Model inference prompt for HotpotQA }}\textcolor{gray}{*/} \\
     Given the context and question, answer the question through step-by-step reasoning based on the context.\\
     ---\\
     Context: \{reference\_docs\} \\
     \\
     Question: \{question\} \\
     --- \\
     \\
     Answer: \\
    \midrule
    \textcolor{gray}{/* }\textcolor{gray}{\textit{ Model inference prompt for GSM8K }}\textcolor{gray}{*/} \\
     Given the following math question, answer it through step-by-step reasoning. \\
     ---\\
     Question: \{question\} \\
     ---\\
     \\
     Answer: \\
    \midrule
    \textcolor{gray}{/* }\textcolor{gray}{\textit{ Unified self-evaluation prompt }}\textcolor{gray}{*/} \\
     Judge the correctness of the answer in the following Q\&A scenario: \\
     \#\#\# \\
     \{model\_response\} \\
     \#\#\# \\
     \\
     Judge: \\
    \bottomrule
  \end{tabular}
\end{table*}
\clearpage




\end{document}